\documentclass[10pt,twocolumn,letterpaper]{article}

\usepackage{cvpr}          

\usepackage{booktabs}
\usepackage{tabularx}
\usepackage{xcolor}
\usepackage{colortbl} 
\usepackage{tikz}
\usepackage{multirow}
\usepackage{pgfplots}
\usepackage{pifont}
\usepackage[most]{tcolorbox}
\usepackage[normalem]{ulem}

\tcbset{
  vbgsbox/.style={
    enhanced,
    boxrule=0pt,
    sharp corners,
    left=1.2mm,right=1.2mm,top=0.9mm,bottom=0.9mm,
    colback=#1,
  }
}

\newcommand{\cmark}{\textcolor{green!60!black}{\ding{51}}}
\newcommand{\xmark}{\textcolor{red!70!black}{\ding{55}}}

\pgfplotsset{compat=1.18}

\usetikzlibrary{arrows.meta,positioning,fit,backgrounds,calc,decorations.pathreplacing}

\definecolor{ELBOBlue}{RGB}{33,41,92}
\definecolor{SSTBlue}{RGB}{6,90,130}

\definecolor{green}{HTML}{009E73}
\definecolor{red}{HTML}{CC0000}

\usepackage{algpseudocode}
\algtext*{EndFor}

\definecolor{cvprblue}{rgb}{0.21,0.49,0.74}
\usepackage[pagebackref,breaklinks,colorlinks,allcolors=cvprblue]{hyperref}
\usepackage{algorithmicx}
\usepackage{algorithm}
\usepackage{algpseudocode}
\usepackage{makecell}
\newcommand{\tsim}{\mathrel{\mkern-6mu\sim\mkern-6mu}}

\def\paperID{6} 
\def\confName{CVPR}
\def\confYear{2026}

\title{Improving Continual Learning for Gaussian Splatting based Environments Reconstruction on Commercial Off-the-Shelf Edge Devices}

\author{Ivan Zaino\\
Politecnico di Torino\\
\and
Matteo Risso\\
Politecnico di Torino\\
\and
Daniele Jahier Pagliari\\
Politecnico di Torino\\
\and
Miguel de Prado\\
VERSES\\
\and
Toon Van de Maele\\
VERSES\\
\and
Alessio Burrello\\
Politecnico di Torino\\
}

\begin{document}
\maketitle
\begin{abstract}
Novel view synthesis (NVS) is increasingly relevant for edge robotics, where compact and incrementally updatable 3D scene models are needed for SLAM, navigation, and inspection under tight memory and latency budgets. Variational Bayesian Gaussian Splatting (VBGS) enables replay-free continual updates for the 3DGS algorithm by maintaining a probabilistic scene model, but its high-precision computations and large intermediate tensors make on-device training impractical.
We present a precision-adaptive optimization framework that enables VBGS training on resource-constrained hardware without altering its variational formulation. We (i) profile VBGS to identify memory/latency hotspots, (ii) fuse memory-dominant kernels to reduce materialized intermediate tensors, and (iii) automatically assign operation-level precisions via a mixed-precision search with bounded relative error.
Across the Blender, Habitat, and Replica datasets, our optimised pipeline reduces peak memory from 9.44\,GB to 1.11\,GB and training time from $\sim$234\,min to $\sim$61\,min on an A5000 GPU, while preserving (and in some cases improving) reconstruction quality of the state-of-the-art VBGS baseline. 
We also enable for the first time NVS training on a commercial embedded platform, the Jetson Orin Nano, reducing per-frame latency by $\sim$$19\times$ compared to 3DGS.
\end{abstract}
    
\section{Introduction}
\label{sec:introduction}

On-device 3D environment reconstruction is a key capability for embodied intelligence, enabling Simultaneous Localization and Mapping (SLAM) and visual navigation under strict compute, memory, and latency constraints. In realistic deployments, the robot will build and refine a scene model from a sequential image stream collected through an on-board camera. However, offloading these data to build an offline model could be detrimental due to bandwidth, connectivity, and privacy constraints~\cite{kehoe2015survey}.

Therefore, a central challenge is to learn scene representations that are both accurate and efficient to be computed on embedded hardware (HW). 
Neural Radiance Fields (NeRF)~\cite{mildenhall2020nerf} approaches provide high-fidelity novel-view synthesis, but are typically too expensive for on-board use due to slow optimisation and limited real-time throughput ($10\!-\!30$ hours on an A5000 GPU~\cite{deng2022depth}). 
3D Gaussian Splatting (3DGS)~\cite{kerbl2023gaussiansplatting} has recently emerged, offering a better quality vs compute efficiency trade-off. 
This has motivated extensive work on \emph{rendering at-the-edge}, including mobile-friendly Gaussian splatting and collaborative/offloaded pipelines~\cite{liu2025voyager,wan2025edge}.

While 3DGS is a promising solution, edge robotics requires not only efficient rendering but also \textbf{on-device training} and model updates to cope with evolving or unknown environments without resorting to data offloading. Sequential updates introduce a continual-learning regime~\cite{wang2024comprehensive}, where naive incremental optimisation can cause catastrophic forgetting and degrade quality in regions that are not revisited. Hence, most 3DGS pipelines rely on replay mechanisms (i.e., storing past views and periodically retraining the model)~\cite{cai2023clnerf,wang2024continualsurvey,matsuki2024gaussiansplattingSLAM,keetha2024splatam}, further increasing compute and memory demands for model re-training. Only a few recent works explicitly target training or incremental updates under resource constraints~\cite{wang2025react3d,wu2024gauspu,li2025rtgs}.

Variational Bayes Gaussian Splatting (VBGS)~\cite{vandemaele2024vbgs} has been designed to cope with the problem of continual updates by casting GS training as variational inference~\cite{blei2017variational} over a probabilistic mixture model. On the other hand, to enable better replay-free continual learning, VBGS relies on high precision (i.e., \texttt{fp64}) and materialises large intermediate tensors during sufficient-statistics accumulation to ensure the numerical stability of the algorithm. In the baseline implementation of~\cite{vandemaele2024vbgs}, VBGS requires $\sim$234 minutes on an A5000 GPU and $9.44$~GB peak GPU memory, which does not fit a typical edge platform memory, such as the $8$~GB available on a Jetson Orin Nano.

In this work, we tackle the problem of porting Novel View Synthesis (NVS) training at the edge, by proposing a precision-adaptive optimisation framework that makes VBGS training feasible on resource-constrained HW without modifying its objective or variational update rules. Our main contributions are:
\begin{itemize}
    \item We profile the baseline VBGS algorithm to localise latency and peak-memory hotspots in the pipeline via function-level profiling.
    \item We propose two optimizations for memory and latency improvement, i.e., (i) fusing memory-dominant kernels to eliminate large intermediate tensors, and (ii) an automatic mixed-precision search assigning operation-level precision under explicit numerical-stability constraints.
    \item We achieve $\sim\!\!4\times$ training speedup and $\sim\!\!9\times$ peak-memory reduction relative to the VBGS baseline~\cite{vandemaele2024vbgs} on an A5000 GPU across the Blender~\cite{mildenhall2020nerf}, Habitat~\cite{savva2019habitat}, and Replica~\cite{straub2019replica} datasets with on-par or better reconstruction error, enabling for the first time its execution on a Jetson Orin Nano. 
\end{itemize}

\section{Background and Related Works}
\label{sec:background}

\subsection{Novel view synthesis}
\label{subsec:novel_view_synt}
\cref{fig:NVM_example} shows a common NVS setup where a set of posed observed images (shown in gray) is used to \textit{train} a model which is then used to \textit{render} novel views from unseen camera poses (shown in red).

A popular approach, NeRF~\cite{mildenhall2020nerf}, effectively models scenes as a neural volumetric radiance field rendered via differentiable volumetric integration, but its ray-marching-style optimization is computationally heavy. Efficient variants~\cite{muller2022instantngp, yu2022plenoxels,chen2022tensorf,fridovichkeil2023kplanes} improve training and/or rendering on desktop GPUs, yet training remains too expensive for edge deployment. For these reasons, we do not further consider NeRF in our analysis, and instead focus on 3DGS~\cite{kerbl2023gaussiansplatting}, the new de facto standard in this field, and on VBGS~\cite{vandemaele2024vbgs}, a probabilistic derivation that improves training in a continual learning setup.

\usetikzlibrary{arrows, decorations.markings}
\tikzset{
vecArrow/.style={
  thick,
  decoration={markings,mark=at position
   1 with {\arrow[scale=2,thin]{open triangle 60}}},
  double distance=1.4pt, shorten >= 10.5pt,
  preaction = {decorate},
  postaction = {draw,line width=1.4pt, white,shorten >= 4.5pt}
  },
}
\definecolor{color1}{RGB}{219,233,246}
\definecolor{color2}{RGB}{228,238,231}
\begin{figure}[t]
\centering
\resizebox{\columnwidth}{!}{
\begin{tikzpicture}[font=\footnotesize, line width=0.8pt]
\node[draw, rounded corners, inner sep=0pt, fill=white] (scene) at (0,0)
{\includegraphics[width=2cm,height=2cm]{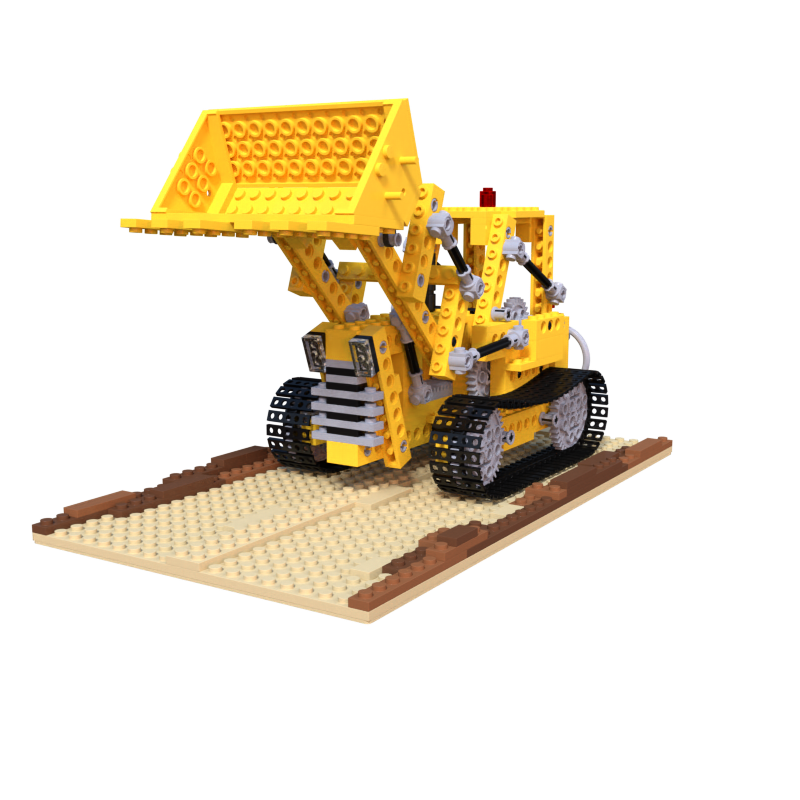}};
\node[draw, rounded corners, minimum width=0.75cm, minimum height=0.45cm, fill=gray!30] (cam1) at (-2.3,1.3) {};
\node[draw, rounded corners, minimum width=0.75cm, minimum height=0.45cm, fill=gray!30] (cam2) at (-2.3,-1.3) {};
\node[draw, rounded corners, minimum width=0.75cm, minimum height=0.45cm, fill=red!10, draw=red] (cam3) at (2.3,1.3) {};
\node[anchor=south] at (cam1.north) {Train cam};
\node[anchor=north] at (cam2.south) {Train cam};
\node[anchor=south, text=red] at (cam3.north) {Test cam};
\draw (cam1.east) -- ($(scene.west)+(0,0.9)$);
\draw (cam1.east) -- ($(scene.west)+(0,0.2)$);
\draw (cam2.east) -- ($(scene.west)+(0,-0.2)$);
\draw (cam2.east) -- ($(scene.west)+(0,-0.9)$);
\draw[red] (cam3.west) -- ($(scene.east)+(0,0.9)$);
\draw[red] (cam3.west) -- ($(scene.east)+(0,0.2)$);
\node[draw, rounded corners, minimum width=2.4cm, minimum height=1.0cm, align=center, fill=color2] (repr) at (7.8,-0.8)
{\textbf{NVS pipeline}};
\node[draw, minimum width=1.2cm, minimum height=1.2cm, fill=gray!30] (I1) at (4.0,1.0) {$I_1$};
\node[draw, minimum width=1.2cm, minimum height=1.2cm, fill=gray!30] (I2) at (5.5,1.0) {$I_2$};
\node[draw=red, minimum width=1.2cm, minimum height=1.2cm, fill=red!10, text=red] (Istar) at (4.75,-0.8) {$I^*$};
\node[anchor=south] (trainlabel) at ($(I1.north)!0.5!(I2.north)$) {\textbf{Training set}};
\node[anchor=south, text=red] (testlabel) at (Istar.north) {\textbf{Test set}};
\begin{pgfonlayer}{background}
  \node[draw, densely dotted, rounded corners, fill=color1, inner sep=8pt,
        fit=(I1)(I2)(Istar)(repr)(trainlabel)(testlabel)] (rightbg) {};
\end{pgfonlayer}
\draw[vecArrow] ($(scene.east)+(0.2,0)$) -- (3,0);
\draw[->, red] ($(repr.west)+(-0.1,0)$) -- ($(Istar.east)+(0,0)$);
\node[anchor=west, text=red] at ($(repr.west)+(-1.05,+0.3)$) {render};
\coordinate (bendTrain) at (repr.north |- I2.east); 
\draw[->] ($(I2.east)+(0.1,0)$) -- node[midway, above] {train} (bendTrain) -- (repr.north);
\end{tikzpicture}
}
\caption{Novel view synthesis setup: observed training views (in gray) are used for training. Test views (in red) are used to evaluate rendering quality from unseen viewpoints.}
\label{fig:NVM_example}
\vspace{-0.5cm}
\end{figure}
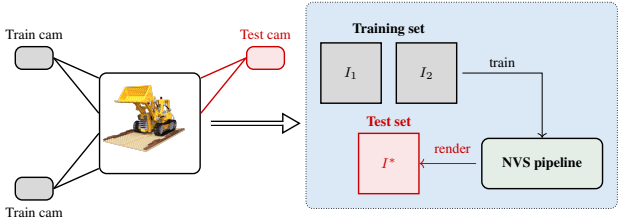

\begin{table}[t]
\centering
\setlength{\tabcolsep}{5.7pt}
\footnotesize
\begin{tabular}{l c c c c c c c}
\toprule
\textbf{Method} &
\textbf{Continual} &
\makecell{\textbf{Replay}\\\textbf{free}} &
\makecell{\textbf{Edge}\\\textbf{training}} &
\makecell{\textbf{COTS HW}} & \\

\midrule

Voyager \cite{liu2025voyager} &

n.a. & n.a. &  n.a. & \cmark \\

ECO-GS \cite{wan2025edge} &

n.a. & n.a. & n.a. & \cmark  \\

3DGS \cite{kerbl2023gaussiansplatting}  &

\xmark & n.a. &  \xmark &\cmark \\



Grendel \cite{zhao2024scaling}&

\xmark & n.a. &  \xmark & \cmark  \\

CLNeRF \cite{cai2023clnerf} &

\cmark & \xmark &  \xmark & \cmark \\

G. S. SLAM \cite{matsuki2024gaussiansplattingSLAM} &

\cmark & \xmark &  \xmark & \cmark \\

SplaTAM \cite{keetha2024splatam} &

\cmark & \xmark &  \xmark & \cmark \\

REACT3D \cite{wang2025react3d} &

\cmark & \xmark &  \cmark & \xmark \\

RTGS \cite{li2025rtgs} & 

\cmark & \xmark & \cmark &\xmark \\

GauSPU \cite{wu2024gauspu} & 

\cmark & \xmark & \cmark &\xmark \\

VBGS \cite{vandemaele2024vbgs} &

\cmark & \cmark &  \xmark &  \cmark \\

\textbf{Ours} &

\cmark & \cmark & \cmark & \cmark \\
\bottomrule
\end{tabular}%

\caption{State-of-the-art GS methods.}
\label{tab:qual_gs_comparison}

\vspace{-0.4cm}

\end{table}

\subsubsection{3D Gaussian Splatting (3DGS)}
\label{subsec:gs_pipeline}
3DGS~\cite{kerbl2023gaussiansplatting} models a scene as a set of anisotropic 3D Gaussians and renders them via differentiable rasterization, achieving real-time rendering with competitive quality. Training optimizes Gaussian parameters (e.g., position, covariance, opacity, and view-dependent appearance) by minimizing a photometric reconstruction loss. 
Follow-up work has largely focused on efficient deployment, especially fast edge/mobile rendering or streaming variants \cite{liu2025voyager,wan2025edge}, but the cost of training and model updates has not been optimized, remaining challenging to execute on Commercial Off-the-Shelf (COTS) edge devices.

This issue becomes central in online mapping and SLAM \cite{matsuki2024gaussiansplattingSLAM, keetha2024splatam,yan2024gsslam}, where sequential observations require frequent updates. The resulting bottleneck has motivated distributed training systems such as Grendel \cite{lu2024grendel} and HW-accelerated or specialized pipelines for faster incremental updates \cite{wang2025react3d,li2025rtgs,wu2024gauspu}. \cref{tab:qual_gs_comparison} shows that 3DGS~\cite{kerbl2023gaussiansplatting} and Grendel~\cite{lu2024grendel} are compatible with COTS HW but are not designed for edge training, being too slow to be executed on board. REACT3D~\cite{wang2025react3d} (and similar approaches~\cite{li2025rtgs,wu2024gauspu}) instead use custom HW to accelerate the update step, thus being incompatible with COTS platforms.

\subsubsection{Variational Bayes Gaussian Splatting}
\label{subsec:vbgs_pipeline}
In on-device NVS, observations arrive sequentially, so models must be updated from a non-stationary stream where catastrophic forgetting can degrade reconstruction in less revisited regions~\cite{mccloskey1989catastrophic}. As shown in~\cref{tab:qual_gs_comparison}, many continual NeRF and GS-based mapping pipelines mitigate this problem by replaying crucial frames seen in the past during model update~\cite{cai2023clnerf,wang2024continualsurvey,matsuki2024gaussiansplattingSLAM,keetha2024splatam}, further increasing the on-device update latency proportionally to the number of replays stored on-board.
To eliminate this latency overhead, which makes edge-training either unfeasible or very slow, VBGS~\cite{vandemaele2024vbgs} treats Gaussian parameters probabilistically and performs closed-form, accumulative variational updates under conjugate exponential-family modelling~\cite{blei2017variational}, which regularize parameter changes and reduce forgetting without the need for replay buffers. However, VBGS requires \texttt{fp64} precision for probabilistic computation, making it difficult to deploy efficiently on resource-constrained edge devices, which motivates the proposed work.

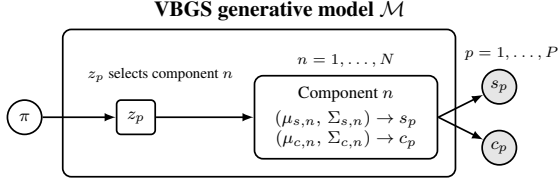
\begin{figure}[t]
\centering
\resizebox{0.9\columnwidth}{!}{%
\begin{tikzpicture}[
  font=\footnotesize,
  >=latex,
  latent/.style={circle,draw,thick,minimum size=16pt,inner sep=0pt},
  obs/.style={latent,fill=gray!20},
  znode/.style={rectangle,draw,thick,rounded corners=2pt,minimum height=16pt,minimum width=18pt,inner sep=1pt},
  box/.style={rectangle,draw,thick,rounded corners,inner sep=5pt,align=center},
  plate/.style={draw,rounded corners,thick,inner sep=8pt},
  arrow/.style={->,thick}
]

\begin{scope}[shift={(0,0)}]
  \node[anchor=west,font=\bfseries] at (-0.2,2.25) {VBGS generative model $\mathcal{M}$};

  \node[latent] (pi) at (-2.2,0.5) {$\pi$};
  \node[znode, right=12mm of pi] (z) {$z_p$};

  \node[box, right=16mm of z, minimum width=30mm] (comp) {Component $n$\\[0.8mm]
    $(\mu_{s,n},\,\Sigma_{s,n}) \rightarrow s_p$\\
    $(\mu_{c,n},\,\Sigma_{c,n}) \rightarrow c_p$
  };
  \node[font=\scriptsize, anchor=south] (text2) at (comp.north) {$n=1,\dots,N$};

  \node[obs] (s) at ($(comp.east)+(1,0.5)$) {$s_p$};
  \node[obs] (c) at ($(comp.east)+(1,-0.5)$) {$c_p$};

  \draw[arrow] (pi) -- (z);
  \draw[arrow] (z) -- (comp);
  \draw[arrow] (comp.east) -- (s);
  \draw[arrow] (comp.east) -- (c);
    \node[font=\scriptsize, align=left] (text) at ($(z.north)+(4mm,4mm)$) {$z_p$ selects component $n$};
  \begin{pgfonlayer}{background}
    \node[plate, fit=(z)(comp)(text)(text2)] (plateN) {};
  \end{pgfonlayer}
  \node[font=\scriptsize] at ($(s.north)+(2mm,2.5mm)$) {$p=1,\dots,P$};

\end{scope}

\end{tikzpicture}
} 
\caption{VBGS generative model $\mathcal{M}$ overview. A point $p$ with position $s_p$ and color $c_p$ is generated by a component $n$ given a latent assignment $z_p$, coming from a categorical distribution with parameter $\pi$. Gray circles denote observed data.}
\vspace{-0.35cm}
\label{fig:VBGS_generative_model}
\end{figure}
VBGS~\cite{vandemaele2024vbgs} models observations with the mixture model $\mathcal{M}$ depicted in \cref{fig:VBGS_generative_model}. $\mathcal{M}$ includes $N$ Gaussian components weighted by $z \tsim \mathrm{Cat}(\pi)$. Each component has two modalities: a spatial variable $s$ (3D Cartesian coordinates) and a color one $c$ (RGB), modelled as multivariate Gaussians conditioned on the components, i.e., $s \!\mid\! z{=}n \tsim \mathcal{N}(\mu_{s,n}, \Sigma_{s,n})$ and $c \!\mid\! z{=}n \tsim \mathcal{N}(\mu_{c,n}, \Sigma_{c,n})$. 
VBGS treats the parameters $\mu$, $\Sigma$ and $z$ as random variables distributed as a Normal–Inverse–Wishart prior on $(\mu,\Sigma) \sim \mathrm{NIW}(m,\kappa,V,n)$ and a Dirichlet prior on the mixture weights, $\pi\,\sim\,\mathrm{Dirichlet}(\alpha)$. These represent the conjugate priors that enable closed-form updates~\cite{vandemaele2024vbgs}.

\begin{algorithm}[t]
\caption{VBGS training pipeline}
\begin{algorithmic}[1]
\Require Initial model $\mathcal{M}_0$, running model $\mathcal{M}$.
\State Initialize stats $(S_{\text{prior}},S_{\text{space}},S_{\text{color}})=S$
\For{each frame $X_t$}

  \Statex
  \begin{tcolorbox}[vbgsbox=blue!6]
  \begin{tcolorbox}[vbgsbox=blue!10]
    $\mathcal{M}_0 \gets \textbf{\textcolor{blue!50}{\texttt{reassign}}}(\mathcal{M}_0, X_t)$:\par\vspace{0.4mm}
    \end{tcolorbox}
    \begin{tabular}{@{}l@{}}
      $\ $ \textbf{for} each batch $X_B$ in $X_t$ \textbf{do} \\
      $\ $ $\ $ $(\mathrm{ELBO},\_) \gets \textbf{\textcolor{ELBOBlue}{\texttt{compute\_elbo\_delta}}}(\mathcal{M}_0, X_t)$\\
      $\ $ \textbf{Sample} $n$ points biased to low $\mathrm{ELBO}$\\
      $\ $ \textbf{Pick} $n$ least-used components\\
      $\ $ \textbf{Reinitialize} selected components in $\mathcal{M}_0$ \\$\ $ \space \space (spatial/color means $\leftarrow$ sampled points)\\
      $\ $ $\textbf{return } M_0$
    \end{tabular}
  \end{tcolorbox}

  \Statex
  \begin{tcolorbox}[vbgsbox=green!6]
    \begin{tcolorbox}[vbgsbox=green!10]
    $(\mathcal{M}, S) \gets \textbf{\textcolor{green!70}{\texttt{fit}}}(\mathcal{M}_0, \mathcal{M}, X_t, S)$:\par\vspace{0.4mm}
    \end{tcolorbox}
    \begin{tabular}{@{}l@{}}
        $\ $ \textbf{for} each batch $X_B$ in $X_t$ \textbf{do} \\
        $\ $ $\ $ $(\_,R) \gets \textbf{\textcolor{ELBOBlue}{\texttt{compute\_elbo\_delta}}}(\mathcal{M}_0, X_B)$ \\
        $\ $ $\ $ \textbf{Compute} ``unsummed" statistics $S_u$\\  $\ $ $\ $ from $X_B$ and $\mathcal{M}$ \\
        $\ $ $\ $ $\Delta S \gets \textbf{\textcolor{SSTBlue}{\texttt{sum\_stats\_over\_samples}}}(R, S_u)$ \\
        $\ $ $\ $ $S \gets S + \Delta S$ \\
        $\ $ $\mathcal{M} \gets \textbf{\texttt{update\_from\_statistics}}(\mathcal{M}, S)$\\
        $\ $ $\textbf{return } (\mathcal{M}, S)$
    \end{tabular}
  \end{tcolorbox}

\EndFor
\end{algorithmic}
\label{alg:vbgs_train}
\end{algorithm}
A high-level view of the training loop of the $\mathcal{M}$ generative model is reported in \cref{alg:vbgs_train}. 
For each image and its corresponding camera parameters, the point cloud $X_t$ (hereinafter denoted as \textit{frame}) is derived. The algorithm starts from the model $\mathcal{M}$ and the sufficient statistics $S$ trained on the previous frame (initialized at 0 for the first frame). 
Additionally, it considers $\mathcal{M}_0$, the initialization model in which the components are reassigned (only the $\mu$ parameters are updated) at every step.
The loop first performs this \textbf{\texttt{reassign}} step to avoid component under-utilization: it evaluates Evidence Lower Bound ($\mathrm{ELBO}$) deltas over batches of points of size $B$, aggregates them, samples $n$ 3D points biased toward low-$\mathrm{ELBO}$ (high-error) regions, randomly samples $n$ Gaussian components among those not yet updated, and reinitializes them by setting their spatial and color means to the sampled points. 
The $\mathrm{ELBO}$ is computed through the \texttt{compute\_elbo\_delta} function, which provides (i) the $\mathrm{ELBO}$, i.e., the variational lower bound on the data log-likelihood, where a lower $\mathrm{ELBO}$ indicates a poorer explanation of the data and (ii) responsibilities $R$, i.e., per-sample soft assignment weights that quantify how strongly each Gaussian component explains each observation. 
The training loop then runs the \textbf{\texttt{fit}} step to update the running model $\mathcal{M}$: for each batch $X_B$, it computes responsibilities $R$, derives per-sample sufficient statistics $S_u$, aggregates them into $\Delta S$ by weighting with $R$ (\texttt{sum\_stats\_over\_samples}), and accumulates them into the global statistics buffer $S$. After processing all batches, VBGS updates the posterior parameters in closed form from $S$ (\texttt{update\_from\_statistics}), returning the updated $(\mathcal{M},S)$.
For more details on the algorithm, readers are referred to \cite{vandemaele2024vbgs}.
\section{Motivation: VBGS bottlenecks}
\label{sec:motivation}
We perform a fine-grained analysis of latency and peak memory consumption across the two macro-routines of VBGS, namely \texttt{reassign} and \texttt{fit} (\cref{alg:vbgs_train}). 
For this analysis, we use a $512\times512$ RGB image with a model size $N$ of $10^5$ components and a batch size $B$ of 500, as in the reference VBGS work \cite{vandemaele2024vbgs}.
\subsection{Latency-driven analysis}
\label{subsec:latency_analysis}
\begin{figure}[t]
    \centering
    \includegraphics[width=\linewidth]{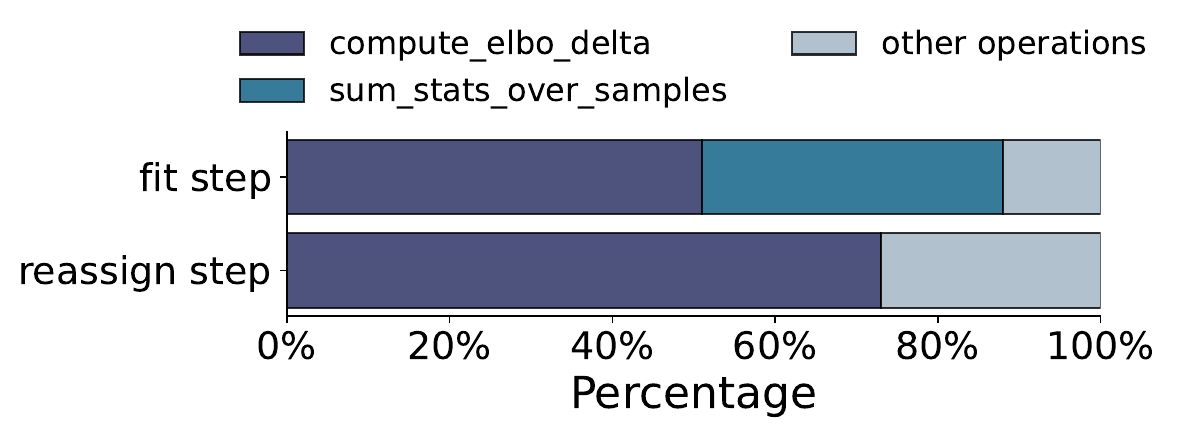}
    \vspace{-0.6cm}
    \caption{Runtime breakdown of VBGS between the reassignment and fitting steps.}
    \label{fig:reassign_fit_split}
\end{figure}
We profile a single iteration of the VBGS training loop using the TensorBoard profiler (XProf) \footnote{\url{https://www.tensorflow.org/guide/profiler}}, which collects traces and attributes runtime to high-level function calls.

\cref{fig:reassign_fit_split} shows the runtime split between the \texttt{fit} and \texttt{reassign} routines.
Two functions dominate the end-to-end latency: the \textbf{\textcolor{ELBOBlue}{\texttt{compute\_elbo\_delta}}}, invoked in both \texttt{reassign} and \texttt{fit}, computes the $\mathrm{ELBO}$ and the data-to-component assignments; the \textbf{\textcolor{SSTBlue}{\texttt{sum\_stats\_over\_samples}}} multiplies the matrix of point-component sufficient statistics $S_u$ with the assignment matrix $R$ and aggregates across the batch dimension $B$ to update model parameters (\cref{alg:vbgs_train}). The cost of both functions is proportional to the batch size $B$ and the number of mixture components $N$.

In the \texttt{reassign} step (\cref{fig:reassign_fit_split}-bottom), the workload is dominated by \textbf{\textcolor{ELBOBlue}{\texttt{compute\_elbo\_delta}}}, occupying 84.4\% of execution time, while the remaining 15.6\% is spent on other operations. In the \texttt{fit} step (\cref{fig:reassign_fit_split}-top), time is split between \textbf{\textcolor{ELBOBlue}{\texttt{compute\_elbo\_delta}}} (42.4\%) and \textbf{\textcolor{SSTBlue}{\texttt{sum\_stats\_over\_samples}}} (45.8\%), with the remaining 11.8\% attributed to other operations. The latter includes updating the model, transposing tensors, or, in the case of the \texttt{reassign} step, modifying unused component means.

\subsection{Memory-driven analysis}
\label{subsec:memory_analysis}

\begin{figure}[t]
    \centering
    \includegraphics[width=\linewidth]{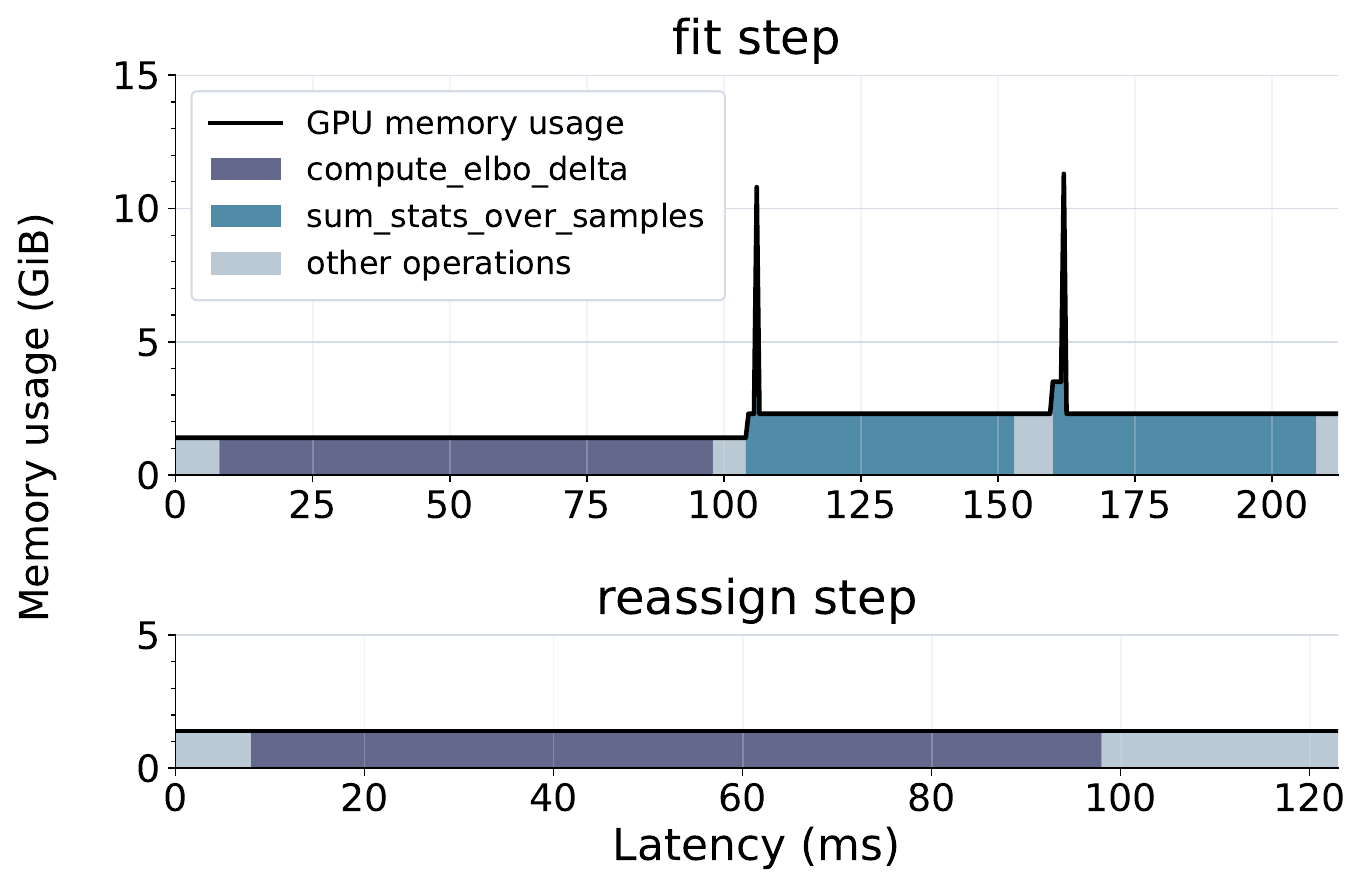}
    \caption{Memory profile during a fit step (top) and a reassign step (bottom) window. The shaded areas represent the function executed during that timestep.}
    \label{fig:memory_spike}
    \vspace{-0.4cm}
\end{figure}

\begin{algorithm}[t]
\caption{\textbf{\textcolor{SSTBlue}{\texttt{sum\_stats\_over\_samples}}} (baseline): materialise-then-reduce weighted statistics.}
\label{alg:sum_stats_over_samples}
\begin{algorithmic}[1]
\Require responsibilities $\mathbf{R}\in\mathbb{R}^{B\times N}$, unsummed statistics $\mathbf{S_u}\in\mathbb{R}^{B\times K}$
\State $B$: batch size; $N$: number of mixture components; $K$: flattened parameter dimension
\For{each parameter $p \in (m,\kappa,V,n)$}
\State $K = \dim(p)$
\vspace{2pt}
\State \textbf{Broadcasted product (materialised)}
\State Initialize ${\mathbf{\Delta S^*}} \gets \mathbf{0}_{B\times N\times K}$
\For{$b \gets 1$ to $B$}
  \For{$n \gets 1$ to $N$}
    \For{$k \gets 1$ to $K$}
      \State \textcolor{red}{$\mathbf{\Delta S^*}[b,n,k] \gets \mathbf{R}[b,n]\cdot \mathbf{S_u}[b,k]$}
    \EndFor
  \EndFor
\EndFor
\State \textbf{Reduction over batch}
\State Initialize ${\mathbf{\Delta S}} \gets \mathbf{0}_{N\times K}$
\For{$n \gets 1$ to $N$}
    \For{$k \gets 1$ to $K$}
        \State ${\mathbf{\Delta S}}[n,k] \gets \sum_{b=1}^{B} \mathbf{\Delta S^*}[b,n,k]$
    \EndFor
\EndFor
\State \Return ${\mathbf{\Delta S}}$
\EndFor
\end{algorithmic}
\end{algorithm}

\cref{fig:memory_spike} shows the memory (and absolute latency) profile of a single iteration of each step in \cref{alg:vbgs_train} on an A5000 GPU (we profile only GPU memory, since we are not constrained by the CPU one). Two transient spikes (the first at 10.9 GB and the second at 11.3 GB) dominate the footprint.
By profiling memory allocations, we found these spikes belonging to \textbf{\textcolor{SSTBlue}{\texttt{sum\_stats\_over\_samples}}} function.
As shown in \cref{alg:sum_stats_over_samples}, this function computes weighted sufficient-statistics reductions using posterior responsibilities $R$ and unsummed statistics $S_u$, which provide the information for updating the parameters for each point of the batch. For each parameter to be updated (color and position prior parameters), this function sums the sufficient statistics over the batch dimension and assigns them to the correct Gaussian component.
In the baseline implementation, this computation follows a \emph{materialise-then-reduce} pattern, resulting in a memory peak during matrix $\Delta S^*$ materialization. 
The biggest allocation happens when updating the parameter $V$ (\cref{subsec:gs_pipeline}, scale matrix of the NIW distribution): this parameter's dimension is $K=3 \times 3 $, which results in an intermediate tensor of shape $(B,N,9)$. With values of $N$ and $B$ parameters specified before, it corresponds to an allocation of $\approx 9.4$~GB in \texttt{fp64}.
This motivates a key optimisation goal: eliminating (or fusing away) large broadcasted intermediates within \textbf{\textcolor{SSTBlue}{\texttt{sum\_stats\_over\_samples}}} to reduce peak memory usage.

\section{Methods}
\label{sec:methods}
\begin{algorithm}[t]
\caption{\textbf{\textcolor{SSTBlue}{\texttt{sum\_stats\_over\_samples}}}: fused contraction (no intermediate allocation)}
\label{alg:sum_stats_over_samples_fused}
\begin{algorithmic}[1]
\Require Rensposibilities $\mathbf{R}\in\mathbb{R}^{B\times N}$, unsummed sufficient statistics $\mathbf{S_u}\in\mathbb{R}^{B\times K}$

\State $B$: batch dim, $N$: number of components, $K$: flattened parameter dimension

\State \textbf{Fused multiply-accumulate contraction}
\State Initialize ${\mathbf{\Delta S}} \gets \mathbf{0}_{N\times K}$

\For{$b \gets 1$ to $B$}
    \For{$n \gets 1$ to $N$}
        \For{$k \gets 1$ to $K$}
            \State \textcolor{green}{${{\Delta S}}[n,k] \gets {\mathbf{\Delta S}}[n,k]
            + \mathbf{R}[b,n]\cdot \mathbf{S_u}[b,k]$}
        \EndFor
    \EndFor
\EndFor

\State \Return ${\mathbf{\Delta S}}$
\end{algorithmic}
\end{algorithm}

\begin{figure}[t]
    \centering
    \includegraphics[width=\linewidth]{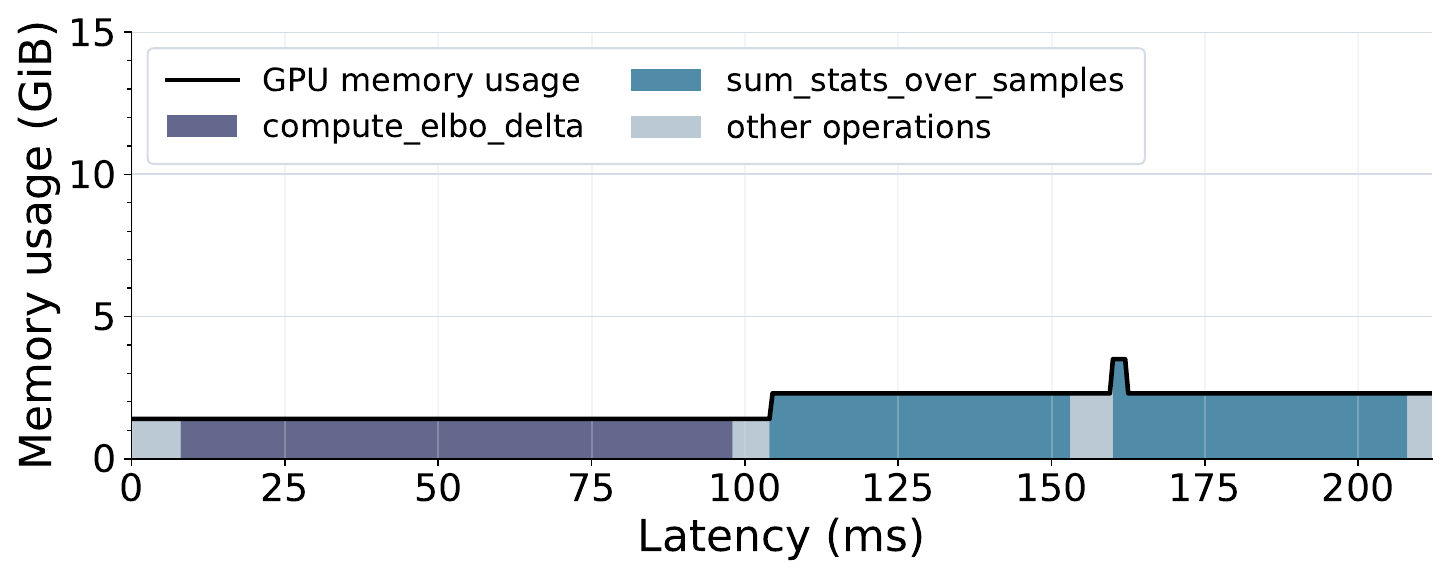}
    \caption{Effect of replacing a materialise-then-reduce implementation with a fused contraction (\cref{alg:sum_stats_over_samples_fused}).}
    \label{fig:memory_spike_fix}
    \vspace{-0.4cm}
\end{figure}
We propose two complementary strategies to make VBGS training feasible on edge platforms.
First, to resolve the observed memory peaks, we applied kernel fusion to the baseline implementation to remove the dominant \emph{transient} allocations identified in \cref{subsec:memory_analysis}.
Second, to jointly reduce runtime and memory footprint, we introduce an automatic mixed-precision pipeline that assigns reduced numerical precision to selected operations while enforcing an explicit numerical-stability constraint.
All the algorithms and techniques discussed hereinafter are based on and implemented in the JAX framework, as the baseline VBGS implementation~\cite {vandemaele2024vbgs}.

\subsection{Memory-driven kernel fusion}
\label{sec:code_modification}
The profiling analysis in \cref{subsec:memory_analysis} shows that peak memory is dominated by
\textbf{\textcolor{SSTBlue}{\texttt{sum\_stats\_over\_samples}}}.

We reformulate it as a fused contraction.
Let $R \in \mathbb{R}^{B\times N}$ denote responsibilities and $S_u \in \mathbb{R}^{B\times K}$ denote the per-parameter unsummed sufficient statistics (flattened), where $K$ is the parameter dimensionality (e.g., $K\!=\!9$ for a $3\times 3$ covariance statistic). The baseline in \cref{alg:sum_stats_over_samples} computes the weighted reduction by first materializing the broadcasted product:
\begin{equation}
\Delta S^{*}[b,n,k] \;=\; R[b,n]\; S_u[b,k], 
\;\: \Delta S^{*}\in\mathbb{R}^{B\times N\times K},
\label{eq:materialize}
\end{equation}
and then reducing over the batch dimension,
\begin{equation}
\Delta S[n,k] \;=\; \sum_{b=1}^{B} \Delta S^{*}[b,n,k], 
\;\: \Delta S\in\mathbb{R}^{N\times K}.
\label{eq:reduce}
\end{equation}
This materialize-then-reduce strategy incurs an additional $O(BNK)$ peak-memory footprint due to $\Delta S^{*}$. We replace it with an equivalent fused contraction (\cref{alg:sum_stats_over_samples_fused}) that directly accumulates along the contracting axis (the batch dimension),

\begin{equation}
\Delta S \;=\; R^{\top} S_u,
\label{eq:fused_contr}
\end{equation}
This reformulation reduces the peak-memory complexity of the weighted reduction from $\mathcal{O}(B\,N\,K)$ to $\mathcal{O}(N\,K)$ (plus the already-required storage of $R$ and $S_u$).
We implement \cref{eq:fused_contr} using a contraction primitive (\cref{alg:sum_stats_over_samples_fused}) so that the JAX-XLA compiler can generate an efficient kernel without materializing broadcasted intermediates.
The effect on memory-over-time behaviour is shown in \cref{fig:memory_spike_fix}. Our new implementation shows no degradation in latency while reducing the memory by 68.9\%.
\subsection{Automatic mixed-precision search}
\label{sec:amp_overview}
The original VBGS algorithm \cite{vandemaele2024vbgs} trains the model with high precision (\texttt{fp64}) to avoid instability in accumulative statistics, normalisation, and likelihood-related computations.
However, it incurs high latency, energy consumption, and memory usage.
We propose an automatic precision-assignment procedure that selectively downcasts operations while preserving numerical correctness to reduce the average precision of VBGS operations.
Our algorithm receives as input a user-defined tolerance $\varepsilon$, a function $f$, and its inputs $x$. Notably, the proposed algorithm is completely general and can be applied to any JAX function, also outside the scope of GS. In the context of this paper, we target the functions \textbf{\textcolor{ELBOBlue}{\texttt{compute\_elbo\_delta}}} and \textbf{\textcolor{SSTBlue}{\texttt{sum\_stats\_over\_samples}}}, that account for the largest share of VBGS's runtime (\cref{subsec:latency_analysis}). The output is a precision configuration $\Pi^*$ for the selected function.

\paragraph{Problem formulation.}
Consider a shape-static function $f$ (i.e., a function whose input, output, and intermediate tensor shapes are fixed and known at compile time) that is staged into a sequence of $L$ elementary operations:
\begin{equation}
f \;=\; f_L \circ f_{L-1} \circ \dots \circ f_1,
\end{equation}
We associate each operation $f_i$ with a precision choice $\pi_i \in \mathcal{P}$, where $\mathcal{P}$ is a finite set of candidate formats (i.e., \{\texttt{fp16}, \texttt{TF32}, \texttt{fp32}, \texttt{fp64}\}).
A mixed-precision configuration is $\Pi=\{\pi_1,\dots,\pi_L\}$, yielding a precision-annotated computation
\begin{equation}
f^{(\Pi)} \;=\; f_L^{(\pi_L)} \circ \dots \circ f_1^{(\pi_1)}.
\end{equation}

We seek a configuration that reduces execution cost while bounding deviation from a high-precision reference.
Let $y^h=f^{(\Pi^h)}(x)$ be the output under the all-highest-precision configuration $\Pi^h$.
For a candidate configuration $\Pi$, we define a numerically robust relative error:
\begin{equation}
\mathrm{err}(\Pi) \;=\; \frac{\|y^h - f^{(\Pi)}(x)\|_2}{\max(\|y^h\|_2,\tau)},
\label{eq:relerr}
\end{equation}
where $\tau>0$ prevents ill-conditioning when $\|y^h\|_2\approx 0$.
The precision search problem is:
\begin{equation}
\Pi^\star \;=\; \arg\min_{\Pi}\; T\!\left(f^{(\Pi)}\right)
\quad \text{s.t.}\quad \mathrm{err}(\Pi) \le \varepsilon,
\label{eq:mp_opt}
\end{equation}
where $T(\cdot)$ denotes measured wall-clock latency and $\varepsilon$ is a user-defined tolerance.

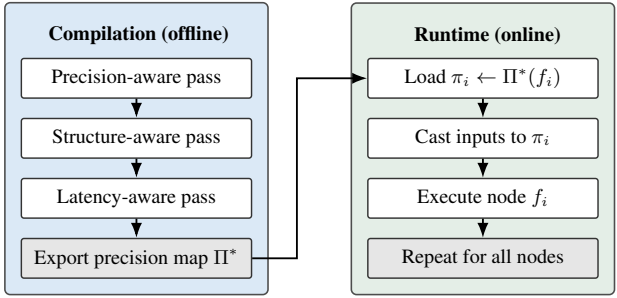
\begin{figure}[t]
    \centering


\newcommand{\FigScale}{0.85}     
\newcommand{\FigFont}{\small} 

\begin{tikzpicture}[
  scale=\FigScale, transform shape,
  font=\FigFont,
  >={Latex[length=2mm]},
  node distance=3mm and 10mm,
  title/.style={font=\bfseries\FigFont, anchor=west},
  stage/.style={
    rounded corners=1pt,
    draw=black!80, line width=0.55pt,
    fill=white,
    align=center,
    minimum width=3.6cm,
    minimum height=6.2mm,
    inner sep=2pt
  },
  emph/.style={stage, fill=black!10},
  group/.style={
    rounded corners=2pt,
    draw=black!70, line width=0.6pt,
    fill=blue!12,
    inner sep=6pt
  },
  arrow/.style={->, line width=0.7pt},
  note/.style={font=\scriptsize, text=black!70, align=center},
]

\definecolor{BoxLeft}{RGB}{219,233,246}  
\definecolor{BoxRight}{RGB}{228,238,231} 
\node[stage] (c1) {Precision-aware pass};
\node[stage, below=of c1] (c2) {Structure-aware pass};
\node[stage, below=of c2] (c3) {Latency-aware pass};
\node[emph,  below=of c3] (c4) {Export precision map $\Pi^{*}$};

\draw[arrow] (c1) -- (c2);
\draw[arrow] (c2) -- (c3);
\draw[arrow] (c3) -- (c4);

\node[title] (ct) at ($(c1.north)+(-1.5,3.5mm)$) {Compilation (offline)};

\node[stage, right=18mm of c1] (r1) {Load $\pi_i \leftarrow \Pi^{*}(f_i)$};
\node[stage, below=of r1]      (r2) {Cast inputs to $\pi_i$};
\node[stage, below=of r2]      (r3) {Execute node $f_i$};
\node[emph,  below=of r3]      (r4) {Repeat for all nodes};

\draw[arrow] (r1) -- (r2);
\draw[arrow] (r2) -- (r3);
\draw[arrow] (r3) -- (r4);

\node[title] (rt) at ($(r1.north)+(-1.2,3.5mm)$) {Runtime (online)};

\begin{pgfonlayer}{background}
  \node[group, fill=BoxLeft,  fit=(ct)(c1)(c4)] (compBox) {};
  \node[group, fill=BoxRight, fit=(rt)(r1)(r4)] (runBox) {};
\end{pgfonlayer}

\coordinate (mid) at ($(c4.east)+(7mm,0)$);
\draw[arrow] (c4.east) -- (mid) |- (r1.west);



\end{tikzpicture}
    \caption{Compilation and runtime behaviour overview of the mixed-precision search algorithm.}
    \label{fig:mixed_search_flow}
\vspace{-0.4cm}
\end{figure}
The search proceeds in three refinement passes (illustrated in \cref{fig:mixed_search_flow}): (i) \textbf{Precision-aware pass:} determines a numerically stable configuration under the constraint in \cref{eq:mp_opt}; (ii) \textbf{Structure-aware pass:} performs local graph exploration to recover safe downcasts missed by the first pass; (iii) \textbf{Latency-aware pass:} removes downcasts that do not yield speedup after accounting for cast overhead.


\subsubsection{Precision-aware pass}
\label{sec:precision_aware_pass}
We represent the staged computation using its jaxpr representation, i.e., the graph-form intermediate representation employed internally by JAX, where nodes correspond to elementary operations and edges encode data dependencies.
For each candidate precision level $\pi\in\mathcal{P}$, we estimate its \emph{global} output sensitivity by downcasting it in isolation and measuring the induced output deviation using \cref{eq:relerr}.
Specifically, let $\Pi^{(i\leftarrow \pi)}$ denote the configuration that matches $\Pi^h$ except that operation $i$ is executed at precision $\pi$.
We define an operation's sensitivity score:
\begin{equation}
M_{\pi}[i] \;=\; \mathrm{err}\!\left(\Pi^{(i\leftarrow \pi)}\right).
\end{equation}
Then, starting from the lowest-precision configuration $\Pi^{\mathrm{low}}$ (all operations downcasted), we iteratively promote operations to the highest precision in order of decreasing sensitivity $M_{\pi}[i]$ of that precision until $\mathrm{err}(\Pi)\le \varepsilon$. We call the newly obtained configuration $\Pi^{1}$.
We then consider \textit{only the nodes that have been upcasted}, and try to set all of them to the 2nd lowest precision. If the error criterion is not met, we repeat the greedy type promotion in reverse sensitivity order described above. We iteratively proceed in the same way for additional intermediate precisions (if any).

\subsubsection{Structure-aware pass}
The configuration obtained after the previous pass, denoted $\Pi^{(pr)}$, is numerically stable but not necessarily cost-optimal.
This can occur because the first pass uses \emph{single-node} perturbations, which may over-promote nodes with high individual sensitivity.
We therefore run a local refinement starting from all downcasted nodes (``seeds''): for each seed, we attempt to downcast its graph neighbours (producers/consumers) by one precision level, keeping the change if the output constraint remains satisfied.
Downcasted nodes become new seeds, expanding the search until no further safe downcasts are found.
The result is a refined stable configuration $\Pi^{(struct)}$.

\subsubsection{Latency-aware pass}
Not all safe downcasts inserted in the previous steps improve wall-clock time, because the cast operation itself can create operational overhead and inhibit compiler node fusion.
Therefore, starting from the lowest precision, we benchmark candidate \emph{cast-bounded regions} (subgraphs executed at identical reduced precision whose inputs/outputs require explicit casts). 
For each region $\mathcal{C}$, we define the net gain:
$
\Delta T(\mathcal{C}) \;=\; T_{\mathrm{high}}(\mathcal{C}) \;-\; \Bigl(T_{\mathrm{low}}(\mathcal{C}) + T_{\mathrm{cast}}(\mathcal{C})\Bigr)
$
where $T_{\mathrm{high}}$ and $T_{\mathrm{low}}$ are the latencies of the region at the benchmarked precision and at the one immediately higher (e.g., \texttt{fp16} and \texttt{TF32}).
If $\Delta T(\mathcal{C})\le 0$, the region is reverted to the next higher precision. If still tagged as a cast-bound region, it is reinserted into the candidates to be benchmarked for the next precision.
This pruning step yields the final configuration $\Pi^\star$, which retains precision reductions that provide latency improvements.

\subsection{Compilation and search overhead}
\label{sec:compilation_overhead}

Our mixed-precision algorithm incurs a one-time offline cost. First, one iteration of VBGS proceeds normally at the highest precision through the first frame to ensure the model is updated and non-zero statistics are computed. When processing frame~2, the optimization is triggered upon the first call to a function marked for compilation (e.g., \textbf{\textcolor{ELBOBlue}{\texttt{compute\_elbo\_delta}}}).
%
%
Notably, in this work, we apply our algorithm to functions whose relative output error is \textit{data independent}, namely \textbf{\textcolor{ELBOBlue}{\texttt{compute\_elbo\_delta}}} and \textbf{\textcolor{SSTBlue}{\texttt{sum\_stats\_over\_samples}}}.
Experimentally, we validated that the precision assignment is independent of scene content, being unchanged for in-domain frames and white-noise, and thus optimized it on a white-noise frame.

\looseness=-1
Let $|V|$ denote the number of operations (nodes) in the staged graph and $|\mathcal{P}|$ the number of candidate precisions. The precision-aware pass requires up to $\mathcal{O}(|V|\cdot|\mathcal{P}|)$ modified evaluations (and corresponding compilations, depending on implementation) to compute sensitivity scores, followed by a linear-time promotion step. The structure-aware and latency-aware passes add a bounded number of evaluations proportional to the attempted local downcasts and pruned regions.

At runtime, the precomputed operation-to-precision map is loaded, and the function is executed through a JAX JIT wrapper, incurring negligible overhead beyond standard JIT compilation. A new offline run is required only if the function code, tolerance $\varepsilon$, candidate set $\mathcal{P}$, or static input shapes/dtypes change. Since VBGS functions are shape-static and environment-agnostic, precision maps are computed once on a workstation and reused on our edge HW for every dataset.
\section{Experimental Results}
\label{sec:results}
\begin{table*}[t]
    \caption{PSNR (dB) on Blender (left block) and Habitat (center block) and Replica (right block).
    All models are evaluated after 200 training frames on 100 unseen validation frames.
    Values are reported as $\mu \pm \sigma$.
    The best performance for each scene is shown in \textbf{bold}.}
    \vspace{-0.2cm}
    \centering
    \footnotesize
    \setlength{\tabcolsep}{2.5pt}
    \begin{tabular}{ccccccccc|ccc|cccc}
        \toprule

        & \textbf{lego} & \textbf{chair} & \textbf{drums} & \textbf{ficus} & \textbf{hotdog} & \textbf{materials} & \textbf{mic} & \textbf{ship}
        & \makecell{\textbf{Van Gogh}\\\textbf{Room}} & \textbf{Apartment} & \makecell{\textbf{Skokloster}\\\textbf{Castle}} 
        & \textbf{Room0} & \textbf{Room1}  & \textbf{Office0} & \textbf{Office1}    \\
        \midrule

        3DGS &
        19.10 & 20.59 & 15.04 & 19.41 & 19.81 & 16.11 & 23.03 & 21.15 &
        20.55 & 19.18 & 14.30 & \textbf{23.76} & \textbf{26.98} & \textbf{33.03} & \textbf{33.88} \\[-1.0ex]
        \small{(w. replay)} &
        \tiny{$\pm 1.02$} & \tiny{$\pm 0.85$} & \tiny{$\pm 1.13$} & \tiny{$\pm 0.90$} &
        \tiny{$\pm 1.86$} & \tiny{$\pm 1.45$} & \tiny{$\pm 0.72$} & \tiny{$\pm 0.82$} &
        \tiny{$\pm 4.02$} & \tiny{$\pm 5.32$} & \tiny{$\pm 4.36$} &
        \tiny{$\pm 4.04$} & \tiny{$\pm 2.71$} & \tiny{$\pm 3.13$} & \tiny{$\pm 2.44$} \\

        3DGS &
        8.86 & 10.73 & 8.25 & 13.22 & 9.72 & 9.66 & 16.14 & 12.27 &
        10.14 & 10.57 & 10.74 & 10.52 & 11.54 & 14.55 & 14.66 \\[-1.0ex]
        \small{(w/o. replay)} &
        \tiny{$\pm 5.79$} & \tiny{$\pm 2.48$} & \tiny{$\pm 5.09$} & \tiny{$\pm 4.63$} &
        \tiny{$\pm 6.10$} & \tiny{$\pm 5.11$} & \tiny{$\pm 3.30$} & \tiny{$\pm 5.34 $} &
        \tiny{$\pm 6.44$} & \tiny{$\pm 6.96$} & \tiny{$\pm 9.45$} &
        \tiny{$\pm 4.73$} & \tiny{$\pm 4.00$} & \tiny{$\pm 11.17$} & \tiny{$\pm 9.59$} \\

        VBGS &
        \textbf{22.17} & 22.44 & 19.50 & 21.93 & 24.32 & 20.67 & 24.53 & 22.28 &
        21.54 & 24.33 & \textbf{17.12} &
        20.74 & 21.60 &27.56 & 30.04 \\[-1.0ex]
        \small{(baseline)} &
        \tiny{$\pm 0.80$} & \tiny{$\pm 1.01$} & \tiny{$\pm 0.50$} & \tiny{$\pm 0.76$} &
        \tiny{$\pm 0.85$} & \tiny{$\pm 1.39$} & \tiny{$\pm 0.61$} & \tiny{$\pm 0.97$} &
        \tiny{$\pm 3.07$} & \tiny{$\pm 3.22$} & \tiny{$\pm 3.22$} &
        \tiny{$\pm 2.65$} & \tiny{$\pm 1.52$} & \tiny{$\pm 2.41$} & \tiny{$\pm 3.23$}\\

        VBGS &
        21.86 & \textbf{22.67} & \textbf{19.66} & \textbf{22.12} &
        \textbf{24.45} & \textbf{20.72} & \textbf{24.75} & \textbf{22.44} &
        \textbf{22.12} & \textbf{24.49} & 17.06 &
        19.79 & 20.52 & 26.97 & 29.39 \\[-1.0ex]
        \small{(optimized)} &
        \tiny{$\pm 0.79$} & \tiny{$\pm 1.05$} & \tiny{$\pm 0.50$} & \tiny{$\pm 0.79$} &
        \tiny{$\pm 0.85$} & \tiny{$\pm 1.43$} & \tiny{$\pm 0.62$} & \tiny{$\pm 0.98$} &
        \tiny{$\pm 2.67$} & \tiny{$\pm 3.22$} & \tiny{$\pm 2.88$} &
        \tiny{$\pm 2.90$} & \tiny{$\pm 1.87$} & \tiny{$\pm 2.39$} & \tiny{$\pm 3.33$}\\
        \bottomrule
    \end{tabular}
    \label{tab:psnr_combined}
    \vspace{-0.5cm}
\end{table*}

\begin{table}[t]
    \centering
    \footnotesize
    \setlength{\arrayrulewidth}{0.3pt} 
    \caption{Performance of 3DGS and VBGS training before/after our optimizations on deployment targets. $^\dagger$OOM on Jetson.}
    \vspace{-0.2cm}
    \begin{tabular}{lcc}
    \toprule
        \rule{0pt}{3ex}  
        \textbf{Implementation} & \makecell{\textbf{Per-frame} \\
        \textbf{average latency}} & \makecell{\textbf{Memory} \\ \textbf{usage}} \\
    \midrule
        A5000 (3DGS w. replay)                & $\sim$410 s          & 4.14 GB          \\
        A5000 (VBGS)                & 70.12 s              & 9.44 GB          \\
        \textbf{A5000 (Optimized VBGS)}  & \textbf{18.33 s}     & \textbf{1.11 GB} \\
        Jetson (3DGS w. replay)               & $\sim$3470 s         & 6.54 GB          \\
        Jetson (Baseline VBGS)           & n.a.$^\dagger$               & n.a.$^\dagger$           \\
        \textbf{Jetson (Optimized VBGS)} & \textbf{$\sim$180 s} & \textbf{2.53 GB} \\
    \bottomrule
    \end{tabular}
    \label{tab:performance_comparison}
    \vspace{-0.5cm}
\end{table}
\subsection{Setup}
\label{sec:experimental_setup}
We evaluate our method on three standard NVS benchmarks, i.e., \textbf{Blender} \cite{mildenhall2020nerf}, \textbf{Replica} \cite{straub2019replica}, and \textbf{Habitat} \cite{savva2019habitat}. Blender contains 8 3D synthetic forward-facing objects, while Replica and Habitat provide, respectively, 4 and 3 photo-realistic indoor environments with accurate camera poses and depth maps. Across datasets, we use the standard train/val splits \cite{vandemaele2024vbgs} and report results on held-out views.

We evaluate task performance as the reconstruction quality measured by the Peak Signal-to-Noise Ratio (PSNR) on novel views. The PSNR is computed as: $\mathrm{PSNR} = 10 \log_{10} \left( \mathrm{255}^2 / {\mathrm{MSE}} \right)$, 
where $\mathrm{MSE}$ denotes the Mean Squared Error, i.e., the average over all RGB channels and all pixels of the squared pixel-wise differences between the reconstructed and ground-truth images.
As previously discussed, we evaluate training performance as the continual \emph{fitting} cost of the GS model, rather than the novel-view \emph{rendering} time emphasized in prior edge-oriented works. We report wall-clock time to fit a single new frame and the peak GPU memory reached during fitting.

We consider two deployment targets: an NVIDIA RTX A5000 workstation GPU (24 GB high-bandwidth memory, 27.8 TFLOPS) and, as an edge target, an NVIDIA Jetson Orin Nano embedded GPU (8 GB CPU and GPU shared memory, 1.28 TFLOPS).
\subsection{Baseline homogeneous-precision results}
\label{sec:baseline_homogeneous}%
\begin{figure}[t]
    \centering
    \includegraphics[width=1.0\linewidth]{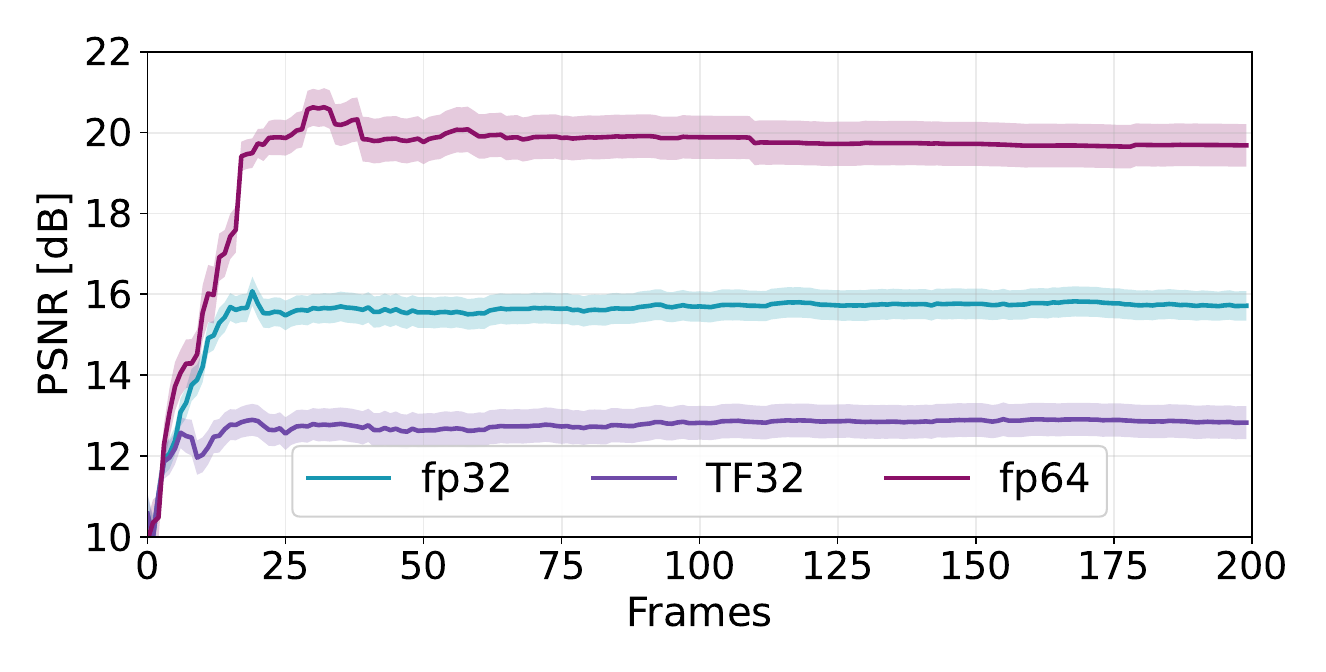}
    \vspace{-0.8cm}
    \caption{PSNR evolution of VBGS under different homogeneous precision variants on the Habitat \emph{Van Gogh Room} environment.}
    \vspace{-0.4cm}
    \label{fig:psnr_ablation}
\end{figure}
Before exploring mixed-precision variants with the algorithm described in Sec. \ref{sec:amp_overview}, we first study homogeneous precision assignments on a representative scenario, i.e., the \emph{Van Gogh Room} from Habitat \cite{savva2019habitat}. 
We select this scene because it represents a middle ground in difficulty within the dataset, i.e., a medium-sized room of average complexity.
We tested \texttt{fp16}, \texttt{fp32}, \texttt{TF32}, and \texttt{fp64} as target precisions.
Starting from the Ampere NVIDIA GPU generation, \texttt{fp32} MatMul can run in TensorFloat-32 (TF32), a 19-bit HW format that reduces mantissa precision to increase throughput: operands are stored in \texttt{fp32}, but the MatMul computation uses \texttt{TF32} and accumulates in \texttt{fp32}.
For brevity, we denote this mode as \texttt{TF32}. 

We fit a VBGS model with $N{=}10^5$ and $B=500$ on 200 frames processed sequentially. After fitting each frame, evaluation is performed by computing the average PSNR and 95\% confidence interval on a fixed set of 100 held-out validation views. 
As shown in \cref{fig:psnr_ablation}, \texttt{fp64} (i.e., the same data type used by VBGS~\cite{vandemaele2024vbgs}) provides the best and most stable reconstruction quality and is used as our baseline.  
When evaluating the PSNR at the end of the fitting, using \texttt{TF32} or \texttt{fp32} instead of \texttt{fp64} reduces the PSNR by 6.85 dB and 3.96 dB, respectively, demonstrating that a homogeneous reduction in precision is detrimental to the algorithm's performance.
Homogeneous \texttt{fp16} results are omitted due to numerical instabilities preventing convergence.

\subsection{Tolerance selection for the mixed-precision search algorithm}
\label{sec:tolerance_selection}
\begin{figure}[t]
    \centering
    \includegraphics[width=0.95\linewidth]{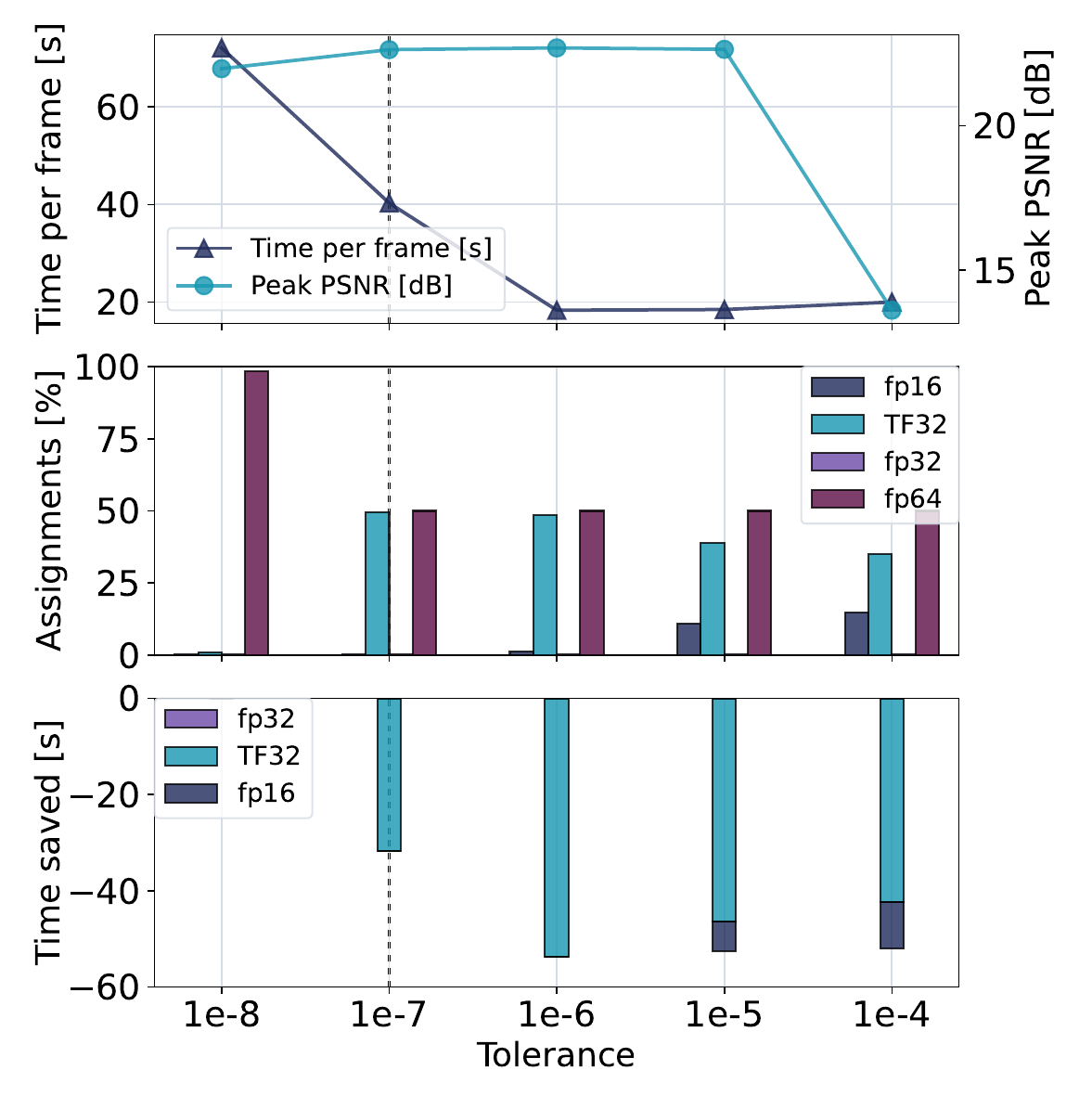}
    \vspace{-0.6cm}
    \caption{Tolerance sweep for the mixed-precision search on Habitat \emph{Van Gogh Room}.}
    \vspace{-0.4cm}
    \label{fig:tolerance_swipe}
\end{figure}

As explained in \cref{sec:amp_overview}, our mixed-precision pipeline requires a user-defined tolerance $\epsilon$.
This tolerance is selected \emph{once} and then kept fixed across all experiments and scenes. 
To choose $\epsilon$, we run a sweep on the same representative workload of the previous \cref{sec:baseline_homogeneous}.
%
%
\cref{fig:tolerance_swipe} summarizes the effect of $\epsilon$ on the runtime-to-quality trade-off and on the resulting precision map.
The top panel reports runtime per frame (left axis) together with peak PSNR (right axis) as $\epsilon$ is relaxed.
$\epsilon{=}10^{-4}$ leads to a clear quality drop, while $\epsilon{=}10^{-5}$ and below preserve stable renderings.
The middle panel explains this behavior through the induced precision assignments: smaller $\epsilon$ values keep most primitives at high precision, whereas larger $\epsilon$ progressively allows lower-precision execution for an increasing fraction of nodes.
Although $\epsilon{=}10^{-7}$ and $\epsilon{=}10^{-6}$ produce approximately the same number of \texttt{TF32} assignments, the $\epsilon{=}10^{-6}$ configuration yields a substantially larger runtime reduction, indicating that the additional \texttt{TF32} assignments occur on particularly runtime-critical operations.
Based on this sweep, we set $\epsilon{=}10^{-6}$ as a single global tolerance for all subsequent experiments, achieving the best quality-to-training-time trade-off. 
Importantly, as discussed in Sec.~\ref{sec:compilation_overhead}, the input frames of the mixed-precision search algorithm are distributed as Gaussian white-noise, thus making the precision assignment environment-agnostic.

\subsection{Benchmark experiments results}

\paragraph{Reconstruction quality.}
\cref{tab:psnr_combined} compares the quality of the trained model in terms of PSNR considering 3DGS (with and without replay buffer), the baseline VBGS, and our optimized VBGS version on the Blender, Habitat, and Replica datasets. Our optimized pipeline yields PSNR that is comparable to the baseline VBGS implementation and outperforms 3DGS without a replay buffer under the same training budget. When compared with the baseline VBGS, improvements are observed in several scenes, whereas small degradations can occur in some cases (e.g., the Sklokoster Castle Habitat, which is the most detailed scene in the datasets and thus requires finer-grained precision to capture details correctly).
Although 3DGS with a replay buffer achieves the highest PSNR on Replica, this setting incurs huge latency overhead as shown in \cref{tab:performance_comparison}. 
Overall, our optimizations are not designed to increase PSNR, but to retain similar reconstruction quality while reducing the system cost of VBGS training. For a qualitative comparison and convergence behavior of VBGS implementations, we refer the reader to the supplementary material.
%
%

\paragraph{Runtime and memory.}
\cref{tab:performance_comparison} reports system-level measurements on a NVIDIA A5000 and on a Nvidia Jetson Orin Nano. On the A5000, our optimized VBGS implementation reduces the per-frame latency from 70.12\,s to 18.33\,s ($4\times$ speedup) and lowers peak GPU memory from 9.44\,GB to 1.11\,GB ($9\times$ reduction) compared to baseline VBGS. This corresponds to reducing the end-to-end training time for a 200-frame run from $\sim$234 minutes to $\sim$61 minutes. Moreover, compared to 3DGS with replay, the end-to-end training time is reduced by more than $22\times$ with our optimized pipeline. On the Jetson Orin Nano, the baseline implementation cannot be executed due to OOM failures, while the optimized pipeline achieves a per-frame latency of 180\,s, reflecting the substantially lower compute throughput and memory bandwidth of the embedded GPU.
The higher memory usage observed on the Jetson compared to the A5000 is due to the Jetson’s unified memory architecture (shared between CPU and GPU), whereas the A5000 reports GPU-only memory usage. 
While slower than the desktop GPU, this result demonstrates the feasibility of VBGS training under the embedded memory budget, which is the primary constraint motivating this work, and although the latency remains high, we argue that in real edge use-cases, input frames could be smaller, and fewer components may need updating. 
These results are dataset-agnostic as the image size, the batch size, and the number of components remain consistent across environments.


\section{Conclusion}
\label{sec:conclusion}

We presented a set of optimizations, including kernel fusion and mixed-precision node assignment, that reduce the peak memory footprint and improve the throughput of VBGS training, enabling on-board novel-view synthesis updates.
Our optimizations reduce the memory footprint compared to the baseline by $\sim$$9\times$ and the average latency by $\sim$$4\times$, while reaching a comparable or better PSNR.
\newpage
{
    \small
    \bibliographystyle{ieeenat_fullname}
    \bibliography{main}
}

\clearpage
\onecolumn
\setcounter{page}{1}

\maketitlesupplementary


\begin{figure}[ht]
    \centering
    \includegraphics[width=1\linewidth]{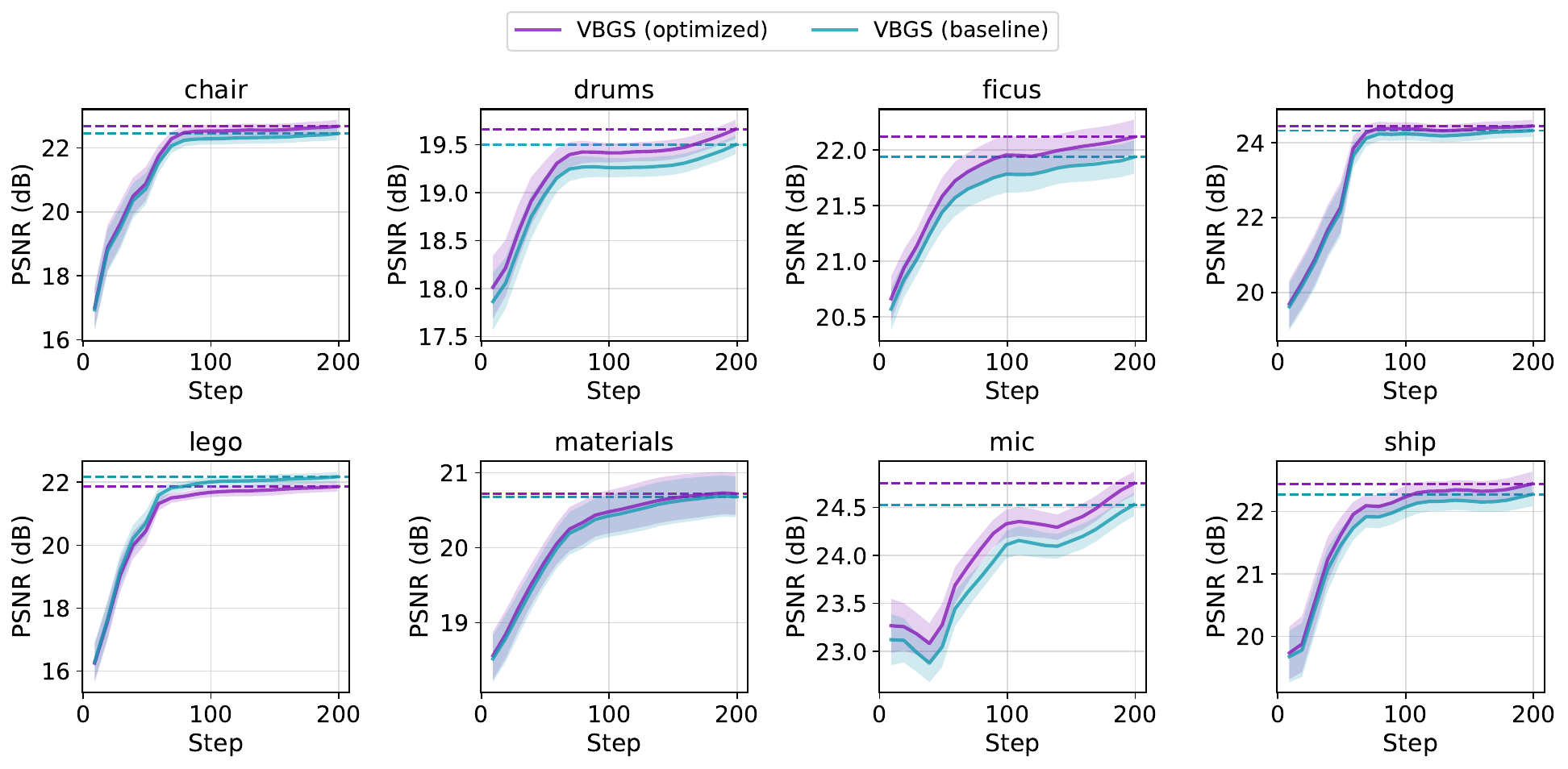}
    \caption{Performance comparison of the VBGS baseline implementation and our optimized training pipeline on 100 validation frames on the Replica dataset.}
    \label{fig:blender_performance_continual}
\end{figure}

\begin{figure}[ht]
    \centering
    \includegraphics[width=0.8\linewidth]{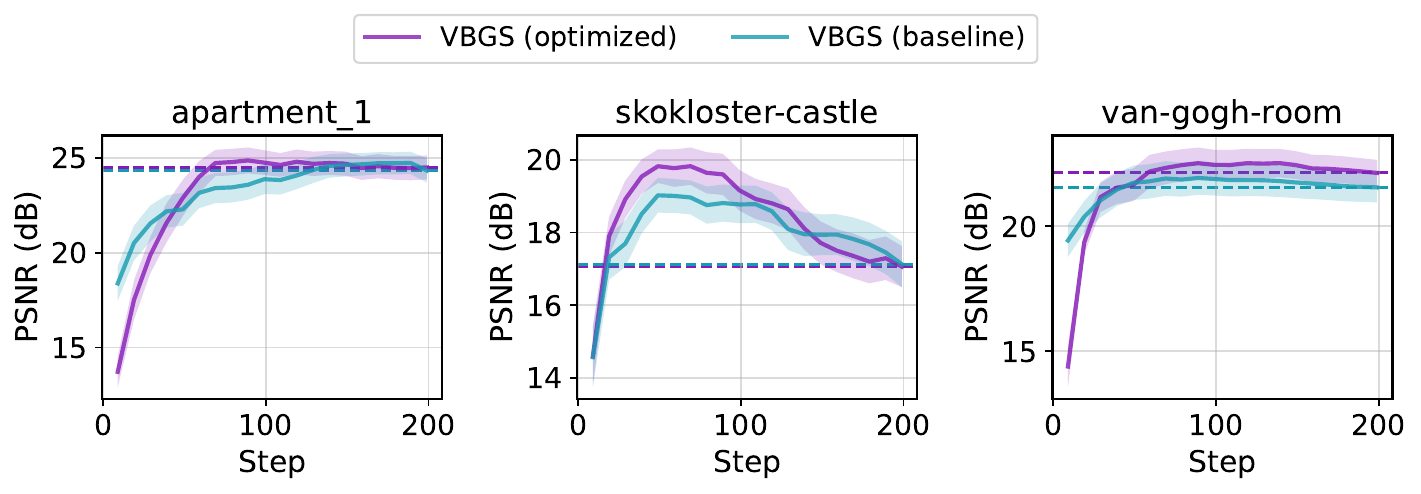}
    \caption{Performance comparison of the VBGS baseline implementation and our optimized training pipeline on 100 validation frames on the Habitat dataset.}
    \label{fig:habitat_performance_continual}
\end{figure}

\begin{figure}[ht]
    \centering
    \includegraphics[width=1.0\linewidth]{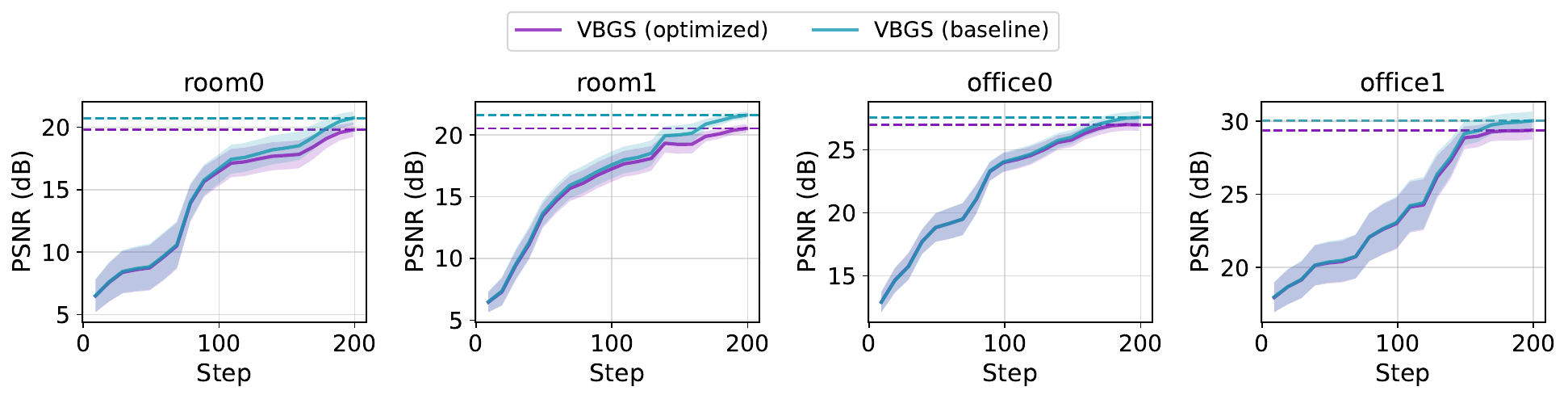}
    \caption{Performance comparison of the VBGS baseline implementation and our optimized training pipeline on 100 validation frames on the Replica dataset.}
    \label{fig:replica_performance_continual}
\end{figure}

\begin{figure}[ht]
    \centering
    \includegraphics[width=1.0\linewidth]{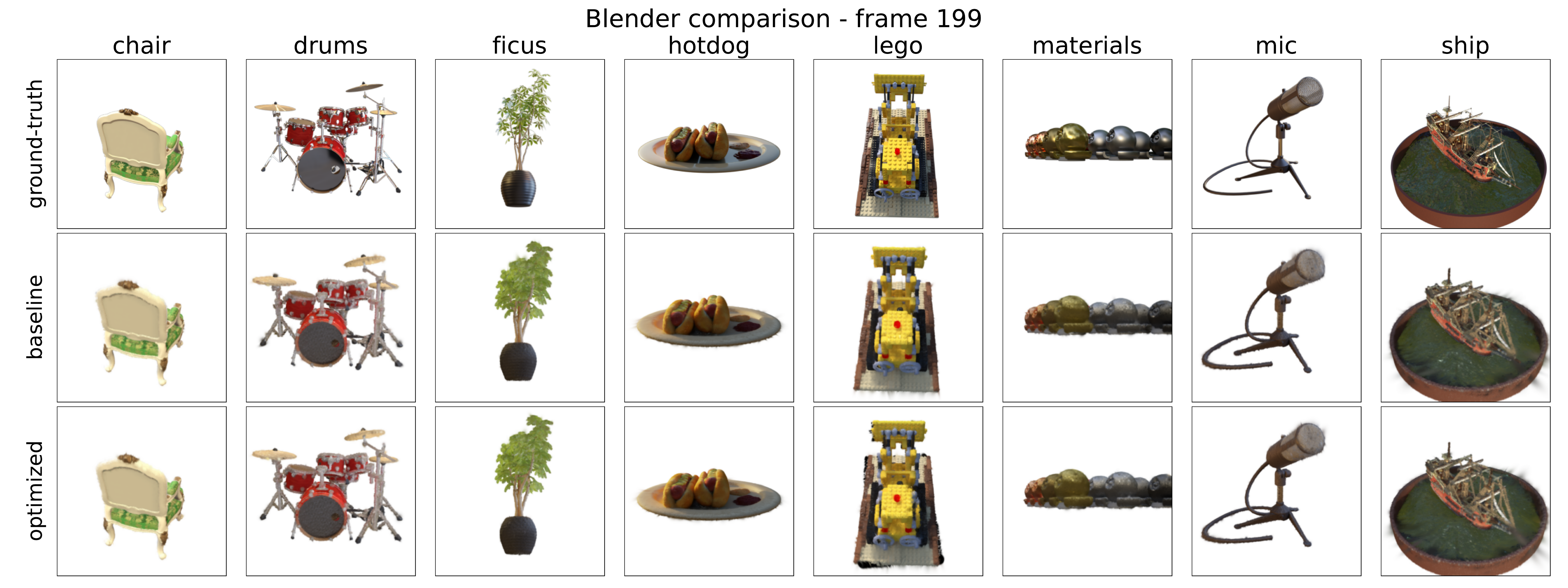}
    \caption{Quality comparison of renderings generated from models trained on 200 Blender frames.}
    \label{fig:blender_quality_comparison}
\end{figure} 

\begin{figure}[ht]
    \centering
    \includegraphics[width=1.0\linewidth]{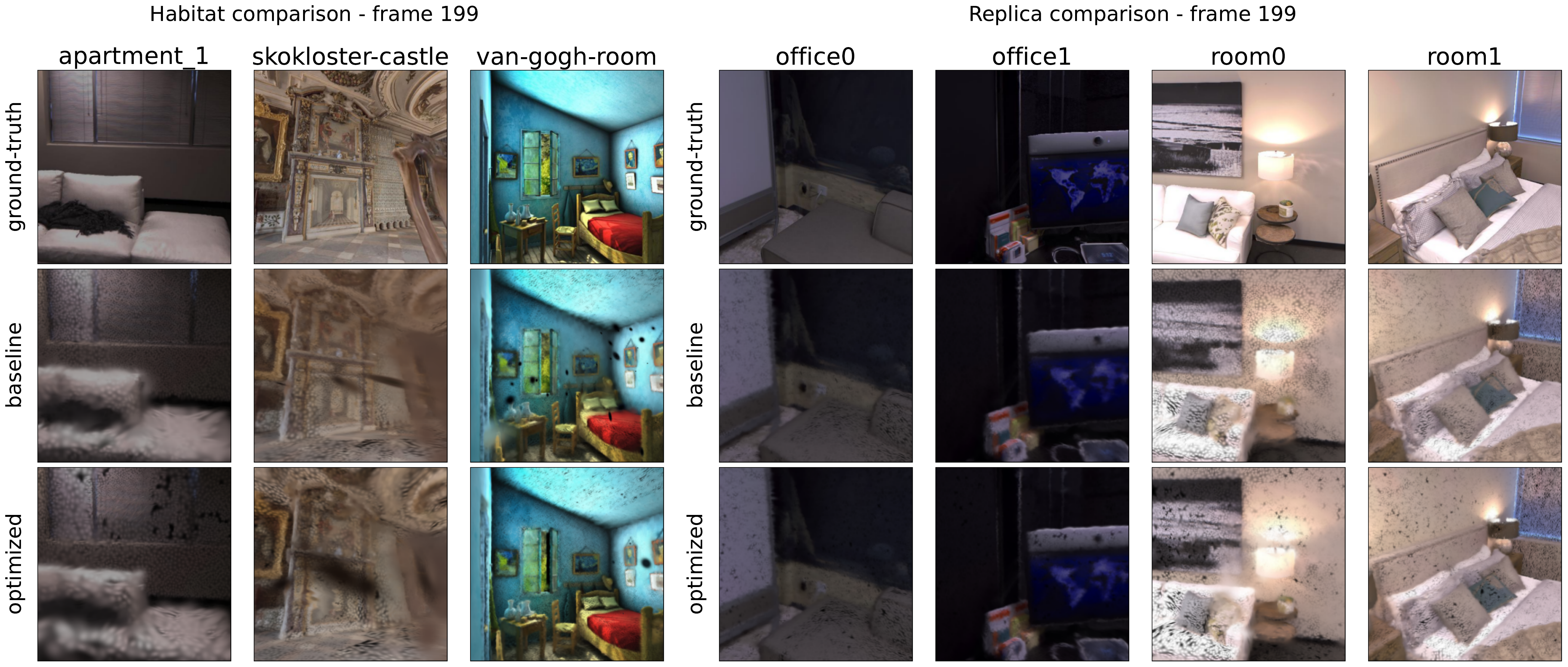}
    \caption{Quality comparison of renderings generated from models trained on 200 Habitat and Replica frames.}
    \label{fig:habitat_replica_quality_comparison}
\end{figure}

\cref{fig:blender_performance_continual}, \cref{fig:replica_performance_continual} and \cref{fig:habitat_performance_continual} provide a finer-grained view of the training timeline of our optimized implementation. The curves indicate that the achieved performance remains stable throughout training. 

\cref{fig:blender_quality_comparison} and \cref{fig:habitat_replica_quality_comparison} report qualitative renderings for the baseline and our optimized pipeline alongside the ground-truth camera image. Overall, both approaches yield comparable visual reconstructions.
\end{document}